\documentclass[letterpaper]{article} 
\usepackage{aaai2027}  
\usepackage[hyphens]{url}  
\usepackage{graphicx} 
\usepackage{natbib}  
\usepackage{caption} 
\usepackage{algorithm}
\usepackage{algorithmic}

\usepackage{newfloat}
\usepackage{listings}
\DeclareCaptionStyle{ruled}{labelfont=normalfont,labelsep=colon,strut=off} 
\floatstyle{ruled}
\newfloat{listing}{tb}{lst}{}
\floatname{listing}{Listing}

\usepackage{booktabs}
\usepackage{threeparttable}
\usepackage{bm}
\title{Automated Analysis Framework for Multilingual Climate-Health Literature Based on Multi-Agent Large Language Model}
\author {
    Yuze Sun\textsuperscript{\rm 1,2},
    Shihui Zhang\textsuperscript{\rm 3}\corresponding,
    Jiancheng Pan\textsuperscript{\rm 1},
    Yunjia Ye\textsuperscript{\rm 4},
    Wentao Luo\textsuperscript{\rm 2},
    Jiahao Li\textsuperscript{\rm 1},
    Quan Zhang\textsuperscript{\rm 5}\corresponding,
    Wenjia Cai\textsuperscript{\rm 1}\corresponding,
    Xiaomeng Huang\textsuperscript{\rm 1}\corresponding
    }

\affiliations{
    \textsuperscript{\rm 1}Department of Earth System Science, Ministry of Education Key Laboratory for Earth System Modelling, Institute for Global Change Studies, Tsinghua University, Beijing, China\\
    \textsuperscript{\rm 2}Huawei Technologies Co., Ltd\\
    \textsuperscript{\rm 3}Renmin University of China, Beijing, China\\
    \textsuperscript{\rm 4}School of Information and AI, Beijing Forestry University, Beijing, China\\
    \textsuperscript{\rm 5}National Institute of Natural Hazards, Ministry of Emergency Management of China, Beijing 100085, China.\\
    
    shihuizhang9414@ruc.edu.cn, quanzhang\_zq@163.com, wcai@tsinghua.edu.cn, hxm@tsinghua.edu.cn

}

\begin{document}

\maketitle

\begin{abstract}
The rapid proliferation of interdisciplinary and multilingual scientific literature has left traditional manual analysis and single-algorithm methods plagued by low efficiency, poor scalability, and insufficient domain adaptability. Targeting the literature analysis needs of the typical interdisciplinary climate–health field, this study proposes a multi-agent large language model automated analysis framework for multilingual scientific literature, which realizes full-process automation covering literature screening, structured information extraction, and standardized integration. With a central coordination module as the core, the framework deploys three dedicated agents for document evaluation, information extraction, and analytical review to mimic the literature analysis thinking of domain experts, and adopts a four-layer hallucination control strategy together with a manual verification procedure to ensure the accuracy and reliability of analytical outcomes. Validated on a bilingual Chinese–English corpus of 32,642 climate-health papers covering China from 1993 to 2023, the framework achieves an F1 score of 0.92 in core information extraction, and completes the extraction and standardization of 2,012 city-literature association pairs, offering effective technical support for large-scale evidence mining in the climate–health research domain.
\end{abstract}

\begin{links}
    \link{Code}{https://github.com/zeaccepted/LLM-based-Al-Agent-for-Climate-Health}
\end{links}

\section{Introduction}
Driven by the rapid development of big data and artificial intelligence, automated scientific literature analysis has become an important research direction in data mining, natural language processing, and knowledge engineering \cite{polak2024extracting, gilligan2023rule}. The continuous expansion of massive academic literature makes traditional literature review methods relying on manual reading, screening, coding, and induction unable to meet the requirements of timeliness, scalability, and standardization \cite{charest2026hermes,ke2025large}. How to automatically, accurately, and stably extract structured knowledge from massive, multilingual, and interdisciplinary unstructured text is the core challenge in current academic big data analysis \cite{xu2024large,adnan2019analytical,adnan2019limitations}.

In recent years, large language models (LLMs) have made breakthroughs in information extraction \cite{dagdelen2024structured}, text classification \cite{sun2023text}, relation mining \cite{zhang2024automated}, and other tasks due to their strong capabilities in semantic understanding \cite{yang2023study}, generalization reasoning \cite{xu2025toward}, and instruction following \cite{zeng2024evaluating}. However, directly applying a single LLM to scientific literature analysis still has significant drawbacks. First, a single model can hardly handle highly specialized multi-task workflows, and is prone to task confusion, unstable output, and non-reproducible results \cite{guo2024large}. Second, LLMs often fabricate entities, relations, and numerical values in professional content generation, leading to severe hallucinations that fail to meet academic rigor \cite{polak2024extracting}. Third, general-purpose LLMs lack domain knowledge, classification systems, geographic rules, and terminological standards, and cannot ensure that extraction results conform to domain specifications \cite{zhang2024comparative}. Fourth, differences in terminological expressions between Chinese and English, cross-disciplinary concept drift, and inconsistent expressions in different journals significantly degrade the structured extraction performance of single models \cite{kleidermacher2026science}.

To address these challenges, multi‑agent LLMs represent a highly promising technical direction \cite{niu2025flow,qian2025scaling}. By decomposing complex workflows into subtasks handled by specialized collaborative and cross‑verifying agents, systems achieve substantial gains in stability, controllability, accuracy and interpretability. While multi‑agent paradigms are well‑studied for dialogue, tool invocation and code generation, dedicated multi‑agent frameworks for academic‑literature analysis remain under‑explored \cite{kumar2026paper}. Existing works lack agent task partitioning aligned with research workflows \cite{zhang2026researchpilot}, hallucination suppression and consistency validation tailored for scientific texts \cite{yang2026inex}, bilingual Chinese‑English standardized term mapping and structured output \cite{zhang2025diting}, as well as end‑to‑end systems validated on real‑world large‑scale literature corpora \cite{go2026lira}. Accordingly, building a modular, verifiable, scalable multilingual multi‑agent literature‑analysis framework carries notable research value for computer science and data mining, and delivers automated knowledge‑mining capacity for diverse interdisciplinary domains \cite{xu2025manalyzer}.

Against the backdrop of converging global climate change and public‑health risks, climate‑health research has emerged as a cross‑cutting frontier spanning environmental science, public health, epidemiology and atmospheric science \cite{romanello20252025}. Over the past three decades, publications on climate change and population health have expanded exponentially, with more than 30,000 relevant papers from mainland China alone \cite{eyzaguirre2025effects,berrang2021systematic}. Released in both Chinese and English, this body of literature features large volume, interdisciplinarity, multilingual coverage and close policy relevance. Such properties render manual curation and classical topic models (e.g., LDA) inadequate \cite{lee2026explainable}, making the domain a suitable test‑bed for multi‑agent frameworks. Systematic collation, structured extraction and evidence‑map construction of these papers carry substantial academic and policy value \cite{berrang2021mapping}. They enable rapid detection of research hotspots, gaps and spatial imbalances to inform research planning \cite{berrang2021systematic,zander2023topic}, while also supporting urban climate adaptation, health‑risk mitigation and public‑health resource allocation \cite{nasari2026healthy}.

The main contributions of this paper are summarized as follows:
\begin{itemize}
\item We present a centralized modular LLM pipeline for bilingual interdisciplinary scientific‑literature analysis. Complex workflows are decomposed into three dedicated prompt‑driven agents (appraiser, extractor, analyst) managed by a central scheduler. This modular design alleviates single‑LLM task overload and output instability, providing a reusable paradigm for domain‑specific literature knowledge mining.

\item We propose a four‑layer hallucination‑control and consistency‑verification pipeline for scientific literature, integrating prompt constraints, cross‑agent logical checking, model‑based confidence scoring, external knowledge‑base filtering, plus expert double‑blind validation. Experiments validate its high‑F1 structured‑extraction performance on the bilingual climate‑health corpus.

\item We implement a full system with multilingual term normalization and three external domain knowledge resources. Evaluated on a large‑scale bilingual climate‑health corpus (>30,000 papers), it generates high‑quality structured datasets and identifies domain research gaps.
\end{itemize}

\section{Method}
\subsection{Overall Design of the Multi-Agent Scientific Literature Analysis Framework}

\begin{figure*}[t]
  \centering
  \includegraphics[width=0.85\textwidth]{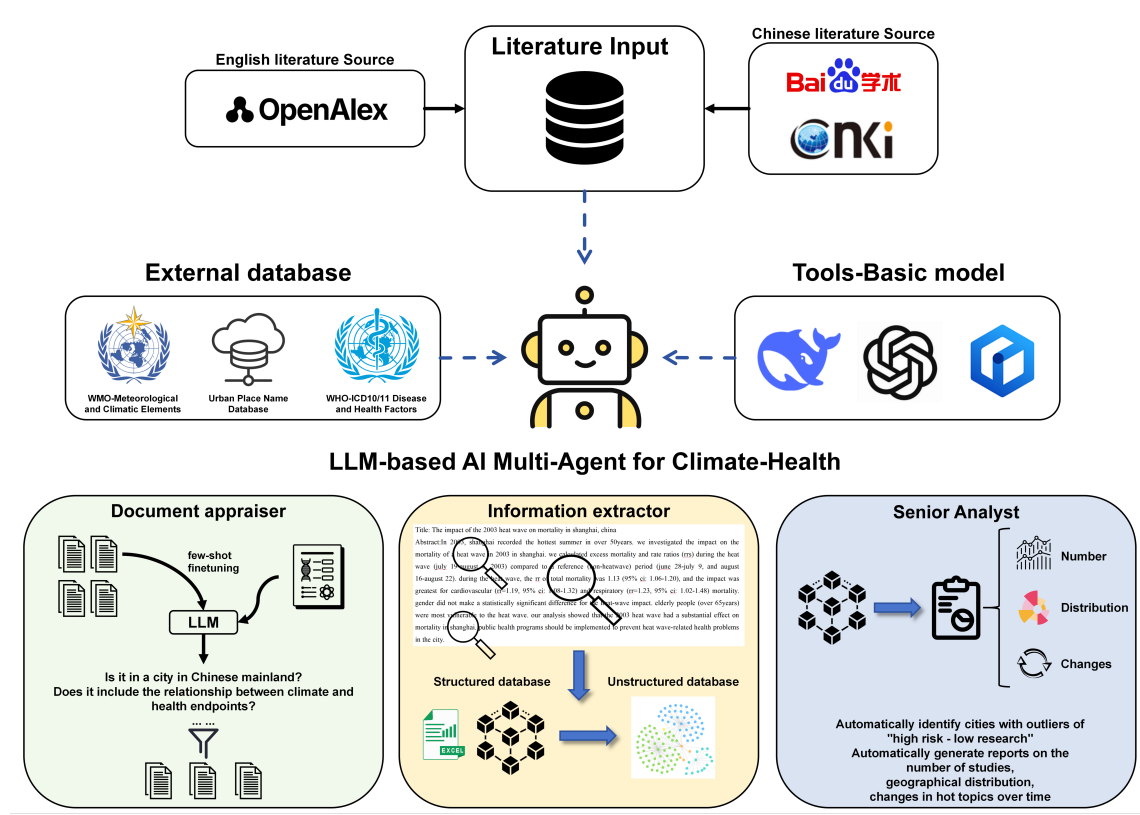}
  \caption{Proposed multi-agent LLM framework for automated climate-health literature analysis. It includes three core agents (Document Appraiser, Information Extractor, and Senior Analyst) supported by external databases and foundational models, enabling end-to-end processing from multi-source literature input to structured data and analysis reports.}
\end{figure*}

The proposed LLM-based climate-health multi-agent analysis framework takes "simulating the collaboration of domain expert teams" as the core design idea, covering the entire process from multilingual literature input to structured knowledge output. The framework adopts a centralized-scheduled multi-agent architecture, with a central coordination module uniformly managing data flow and execution order, and three functionally dedicated analysis agents: document appraiser agent, information extractor agent, and senior analyst agent.

Starting from multilingual literature inputs, our dataset is compiled from OpenAlex for English‑language publications \cite{openalex}, as well as CNKI and Baidu Scholar for Chinese‑language literature \cite{cnki,baiduscholar}. For each retrieved document, we uniformly extract titles, abstracts and keywords while discarding non‑text components such as figures and mathematical formulas, yielding standardized text inputs for our multi‑agent pipeline. To furnish domain‑specific knowledge and computational capabilities for agent modules, we further incorporate external domain databases, including the WMO meteorological element library \cite{wmo_metelib}, the Chinese urban gazetteer \cite{cn_placelib}, and the WHO disease‑and‑health‑factor repository \cite{who_icd}, alongside commercial large‑language‑model APIs (DeepSeek, ChatGPT, Ernie Bot, etc.). Note that these LLMs serve as general‑purpose reasoning backbones, and no fine‑tuning or version‑specific optimization is performed in our pipeline. The pipeline produces standardized structured outputs covering research cities, climate factors, health outcomes, population attributes, and other semantic fields, together with analytical feature reports. The overall architecture and execution workflow of the full system are depicted in Figure 1.

\subsection{Evaluation: Document Appraiser Agent}

The document appraiser agent acts as the initial screening gatekeeper, aiming to quickly identify valid literature meeting the research scope from large-scale raw literature. Based on preset domain inclusion criteria, the agent judges whether literature contains key information such as climate-related exposure, health outcomes, and Chinese city-level research locations, and outputs a binary classification result of "relevant/irrelevant".

To ensure screening consistency, the agent only makes judgments based on explicit information in titles, abstracts, and keywords without additional reasoning or extension. In prompt design, domain specifications and a small number of annotated samples are used to constrain the agent, enabling stable handling of expression differences between Chinese and English and maintaining cross-language screening consistency. This step significantly reduces computational overhead in subsequent extraction stages and is the foundation for efficient operation of the entire pipeline.

\subsection{Extraction: Information Extractor Agent}
As the core functional module of the framework, the information extractor agent is responsible for extracting standardized and quantifiable key information from qualified literature. Extracted content includes research city, climate exposure factor, health outcome, vulnerable population, socio-economic influencing factors, etc. All outputs are strictly aligned with the unified domain classification system to ensure standardized, statistically comparable terminology. For multi-valued fields such as multiple cities, climate factors, or health outcomes mentioned in a single paper, the agent normalizes them into atomic, independent records to ensure data usability, and outputs standardized, ready-to-analyze structured and unstructured datasets after cleaning and integration. Each extracted entity is attached with a model‑estimated confidence score (1‑5 scale), elicited via prompt instruction. The score reflects the LLM’s self‑assessment on whether the target entity is explicitly supported in source text and matches domain terminology resources. In our pipeline, records with confidence score below threshold \(T=3\) are filtered out in the later analyst‑agent stage.

To avoid hallucinations and fabricated content, the extraction process is strictly restricted to extractive tasks: the model can only extract information explicitly stated in the original text without supplementing, inferring, or generating unmentioned content. When atomizing multi‑value outputs (multiple cities / climate‑health pairs), the agent is constrained to only pair entities that are co‑occurring in the same original sentence fragment; arbitrary cross‑combination across disjoint text spans is strictly prohibited via prompt constraints. Each extracted field is accompanied by a confidence score calculated by the model for subsequent verification and filtering. The agent supports automatic Chinese-English term mapping, unifying professional nouns with different languages and expressions into standard categories to achieve aligned analysis of multilingual literature.

\subsection{Analysis: Senior Analyst Agent}
The senior analyst agent undertakes quality control, result integration, and automatic feature report generation, serving as both the final quality gate and the value-added analysis module of the pipeline. It performs multi-dimensional verification on the structured output of the information extraction agent, including field completeness checks, terminology standardization against domain ontologies, logical consistency validation such as common-sense checks between climate exposures and health outcomes and validation of city names and administrative divisions, and confidence-based filtering to eliminate low-confidence and hallucinated entries. For standardized domain classification alignment, we adopt a hybrid mechanism: core category ontology is embedded in system prompts, meanwhile the agent invokes external domain resources as lookup tool calls. The three external domain resources (WMO meteorological element library, Chinese city place‑name library, WHO‑ICD disease library) are wrapped as fuzzy/exact‑match lookup tools. Given an extracted entity string, the analyst agent triggers tool calls to normalize raw terms to standard ontology entries, filter invalid geographic names, and validate health‑outcome terminology. Knowledge bases are not fully stuffed into prompt context and are only accessed during verification phase.

Based on validated data, this tool auto-generates analysis reports. It includes regional and thematic research statistics, geographic distribution patterns, temporal trend shifts of research hotspots, and outlier detection. It also identifies high-risk yet understudied areas to point out research gaps and guide future studies.

\section{Experiments}

\subsection{Experimental Settings}
We comprehensively evaluate the framework from four dimensions: extraction accuracy, operational efficiency, multilingual capability and practical domain value. For extraction performance, we adopt Precision, Recall and F1-score to assess three core extraction tasks (geographic location, climate exposure factor, health outcome), with manually annotated gold labels from two domain experts.

Our experiments rely on a bilingual Chinese–English climate‑health literature corpus covering mainland China (1993–2023). Raw papers are retrieved from OpenAlex, CNKI and Baidu Scholar, followed by deduplication, metadata normalization and text structuring, yielding a final corpus of 32,642 documents.
The multi-agent pipeline is built on Python 3.9 and Prefect for centralized workflow scheduling, with three functional agents connected to mainstream LLMs. Three external domain knowledge bases are incorporated to standardize extracted entities. We randomly select 800 bilingual papers for independent expert annotation, achieving an inter-annotator agreement of 0.87 after resolving conflicts.

Besides extraction metrics, we further measure system throughput for efficiency comparison, compute task F1-scores separately on Chinese and English texts for multilingual evaluation, and adopt effective extraction rate and data standardization rate to quantify practical application value.
Detailed metric definitions, corpus construction procedures, experimental configurations, complete agent prompts and annotation protocols are provided in \textbf{Appendix}.

\subsection{Experimental Results and Analysis}

\subsubsection{Core Information Extraction Accuracy}
Table~\ref{tab:Extraction Performance} reports results for our three core extraction tasks, with an overall F1‑score of 0.92, validating the multi‑agent architecture for rigorous scientific‑literature analysis. Geographic extraction yields the highest F1 (0.94), aided by standardized city names and external place‑name database validation. Climate‑exposure extraction achieves F1 = 0.91, constrained by inconsistent terminology in source papers but improved by the analyst‑module normalization. Health‑outcome extraction reaches F1 = 0.92, performing well for prevalent diseases with slight recall degradation for rare health outcomes. Higher overall precision (0.93) than recall (0.92) confirms the efficacy of our four‑layer hallucination‑control strategy, yielding a false‑positive rate of merely 7\% and achieving low‑hallucination, high‑precision extraction.

To further investigate the advantages of our modular agent pipeline, we compare against a suite of representative baselines including single‑LLM, domain‑trained BERT‑NER, AutoGen multi‑agent implementation and existing open‑source literature‑analysis systems. As shown in Table~\ref{tab:baseline}, straightforward single‑prompt LLM yields relatively lower F1‑score around 0.82. Domain‑specific BERT‑NER and existing multi‑agent literature tools obtain moderate improvements, while our proposed pipeline achieves the best overall precision, recall and F1‑score, demonstrating the benefit of task‑specific agent decomposition together with multi‑layer hallucination mitigation.

We conduct component‑wise ablation experiments to quantify the contribution of each key module. Table~\ref{tab:ablation_hallucination} reports results when removing one hallucination‑control or orchestration component at a time. All variants exhibit clear performance drops compared with the full system. Removing cross‑agent verification or external knowledge‑base filtering leads to obvious precision degradation, while disabling confidence‑score filtering increases false positives and hurts precision. Even though the central coordination module only provides deterministic workflow scheduling, removing it degrades overall performance due to broken data flow and inconsistent intermediate outputs across processing stages.

We also study the effect of different task‑decomposition granularities via agent‑number ablation in Table~\ref{tab:ablation_agent}. The single‑agent setting, which squeezes all screening, extraction and verification work into one LLM call, suffers from task overload and yields the worst F1‑score. Merging two stages into a single agent partially recovers performance. Our three‑agent decomposed architecture obtains the highest F1, verifying that splitting complex literature‑analysis workflows into specialized sub‑tasks effectively stabilizes LLM outputs and improves extraction quality.

We further verify the effectiveness of model‑estimated confidence scoring. We split test predictions by confidence threshold $T=3$: high‑confidence samples ($score\ge 3$) achieve F1=0.94, while low‑confidence samples ($score<3$) only achieve F1=0.66. This confirms that the model‑generated confidence score is positively correlated with actual extraction accuracy and serves as a useful filter for hallucinated entries.

\begin{figure*}[t]
  \centering
  \includegraphics[width=0.9\textwidth]{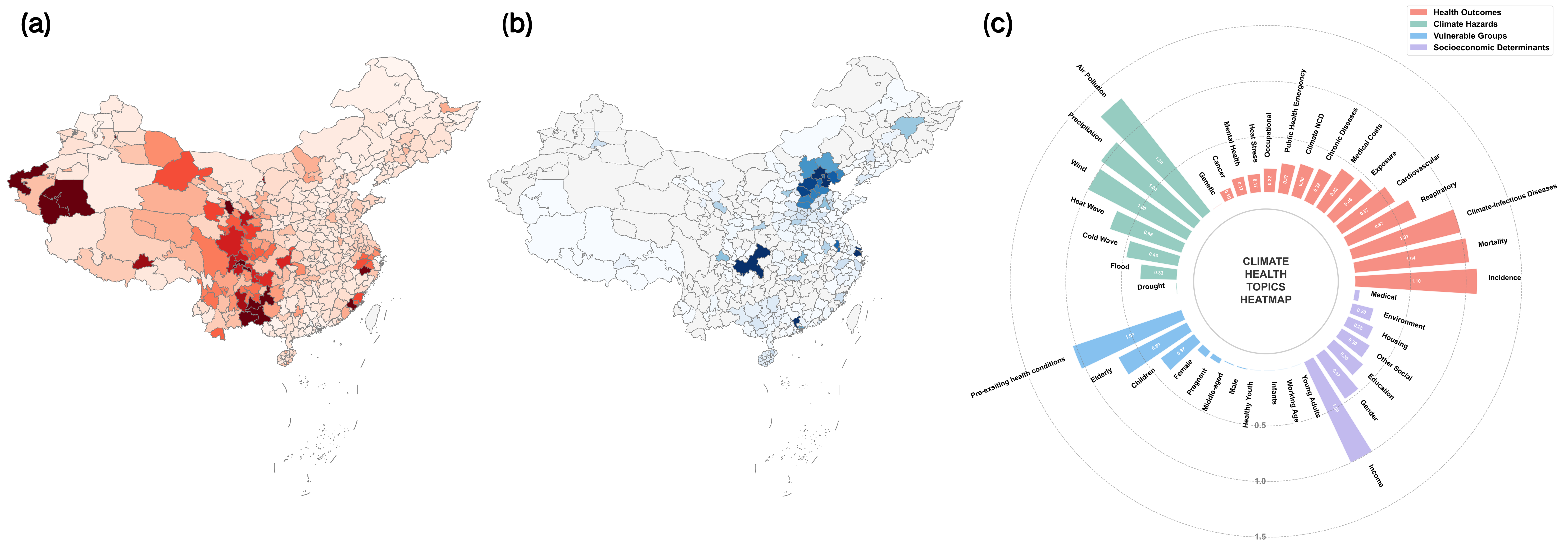}
  \caption{Overview of spatial and thematic patterns in China’s climate-health research. (a) City-level actual risk growth rate across China. \cite{cai20252025} (b) Research hotspots automatically identified by the Senior Analyst framework. (c) Circular heatmap of climate-health research topics, illustrating the relative attention to four categories: health outcomes, climate hazards, vulnerable groups, and socioeconomic determinants, as automatically summarized by the Senior Analyst.}
\end{figure*}

\begin{table*}[t]
\begin{center}
  \caption{Performance Comparison of the Proposed Framework on Core Information Extraction Tasks}
  \label{tab:Extraction Performance}
  \begin{tabular}{cccc}
    \toprule
    Extraction Task&	Precision (P)$\uparrow$&	Recall (R)$\uparrow$&	F1-Score (F1)$\uparrow$\\
    \midrule
    Geographic Information (City)&	0.95&	0.93&	0.94\\
    Climate Exposure Factor&	0.92&	0.91&	0.91\\
    Health Outcome&	0.93&	0.92&	0.92\\
    Overall Average&	0.93&	0.92&	0.92\\
  \bottomrule
\end{tabular}
\end{center}
\end{table*}

\begin{table*}[t]
\centering
\caption{Baseline comparison on three core extraction tasks and overall performance}
\label{tab:baseline}
\begin{tabular}{lccc}
\toprule
Method &	Precision (P)$\uparrow$&	Recall (R)$\uparrow$&	F1-Score (F1)$\uparrow$\\
\midrule
Single LLM (full merged prompt) & 0.84$\pm$0.02 & 0.81$\pm$0.03 & 0.82$\pm$0.02 \\
BERT‑based domain NER \cite{devlin2019bert}& 0.86$\pm$0.03 & 0.83$\pm$0.02 & 0.84$\pm$0.01 \\
AutoGen (sequential multi‑agent) \cite{wu2023autogen}& 0.88$\pm$0.02 & 0.86$\pm$0.02 & 0.87$\pm$0.02 \\
LiRA (Literature Review Agents) \cite{go2026lira}& 0.89$\pm$0.01 & 0.87$\pm$0.02 & 0.88$\pm$0.03 \\
Paper Circle \cite{kumar2026paper}& 0.91$\pm$0.02 & 0.90$\pm$0.01 & 0.90$\pm$0.02 \\
\midrule
Ours (full modular agent pipeline) & \textbf{0.93$\pm$0.01} & \textbf{0.92$\pm$0.01} & \textbf{0.92$\pm$0.01} \\
\bottomrule
\end{tabular}
\begin{tablenotes}
\item Note: All baseline models are re‑implemented and evaluated on our 800‑paper gold‑standard corpus for fair comparison. We finetune hyper‑parameters for each baseline to achieve their best performance on our dataset
\end{tablenotes}
\end{table*}

\begin{table*}[h]
\centering
\caption{Ablation study of agent architectures with different task‑decomposition granularities}
\label{tab:ablation_agent}
\begin{tabular}{lccc}
\toprule
Variant & Precision (P)$\uparrow$&	Recall (R)$\uparrow$&	F1-Score (F1)$\uparrow$\\
\midrule
Single‑agent (all tasks in one agent) & 0.84$\pm$0.02 & 0.81$\pm$0.03 & 0.83$\pm$0.02 \\
Two‑agent (appraiser + merged extractor‑analyst) & 0.89$\pm$0.02 & 0.88$\pm$0.02 & 0.88$\pm$0.02 \\
Three‑agent (our full pipeline) & \textbf{0.93$\pm$0.01} & \textbf{0.92$\pm$0.01} & \textbf{0.92$\pm$0.01} \\
\bottomrule
\end{tabular}
\end{table*}

\subsubsection{Multilingual Processing Performance}
The framework exhibits excellent processing performance for both Chinese and English bilingual literature, with an overall F1-score of 0.93 for English literature and 0.91 for Chinese literature, showing a small gap between the two and good cross-language adaptability. The slightly lower accuracy of Chinese literature is mainly due to non-standard terminological expressions in some Chinese journals, but after standardization correction and manual verification, the accuracy remains stable above 0.91. The framework successfully extracts 92.1\% of city-literature association pairs from Chinese literature, compensating for the evidence bias problem in traditional English single-database analysis.

\subsubsection{Practical Domain Application Effects}
Applied to 32,642 papers, the framework successfully extracts and standardizes 2,012 city-literature association pairs and generates high-quality structured datasets:

\begin{itemize}
\item {\texttt{Effective extraction rate}}: Among 1151 relevant papers, 1128 papers successfully extract core information, with an effective extraction rate of 98\%.
\item {\texttt{Data standardization rate}}: The standardization rate of extraction results reaches 99\%, with all climate factors and health outcomes matching the domain classification framework.
\item {\texttt{Research gap identification}}: Based on the dataset, the framework clearly identifies spatial imbalance (concentrated research in eastern coastal cities, insufficient research in central and western regions, Gini coefficient 0.761) and thematic gaps (focus on exposure-disease relationship, inadequate studies on vulnerable populations and socio-economic factors) in China's climate-health research. The relevant results are visualized in Figure 2, offering explicit guidance for future research in this field.
\end{itemize}

Experimental results demonstrate that the proposed multi-agent LLM framework performs excellently in information extraction accuracy, system efficiency, multilingual processing capability, and domain application value. It achieves a core information extraction F1-score of 0.92, yields substantial efficiency gains relative to fully manual literature analysis, and successfully realizes automated analysis and knowledge mining of large-scale multilingual climate-health literature, with clear academic value and practical application prospects.

\section{Conclusion and Discussion}
This work builds a multi‑agent LLM framework for multilingual interdisciplinary scientific literature and systematically validates the efficacy of agent collaboration for automated literature mining. Experiments demonstrate that our three‑stage agent architecture (document evaluation — information extraction — advanced analysis) substantially boosts extraction accuracy, mitigates hallucinations and improves system stability over single‑LLM baselines. This framework delivers an efficient evidence‑synthesis tool for climate‑health research and provides a reusable multi‑agent paradigm for literature‑oriented data mining and NLP. As academic literature keeps expanding, automated, standardized, low‑hallucination knowledge extraction will grow increasingly critical. Our approach is generalizable to diverse interdisciplinary literature‑analysis scenarios, helping orient research resources toward high‑priority topics and advancing the integration of scientific discovery and decision‑making support.

Despite its promising accuracy and efficiency, the framework still has several limitations:
\begin{itemize}
\item The system currently only analyzes based on titles, abstracts, and keywords without deep extraction of full-text content, which may miss more detailed information such as research design, sample size, and statistical models.
\item Although hallucinations have been greatly reduced, a small number of extraction errors may still occur in literature with vague expressions, missing information, or non-standard language, requiring manual spot checks as the final guarantee.
\item The system relies on external LLM interfaces and may be limited by call rates in extremely large-scale literature scenarios. Localized open-source models can be adopted in the future to further improve autonomy and controllability.
\end{itemize}

Based on the above findings and limitations, future expansion can be carried out in multiple directions:
\begin{itemize}
\item Extend the framework to full-text literature extraction, combined with PDF parsing, chapter segmentation, and table understanding to extract complete research methods, data sources, and statistical results.
\item Introduce dynamic knowledge updates to monitor newly published literature and incrementally refine structured databases for real-time domain trend tracking.
\item Generalize the multi-agent architecture to broader interdisciplinary scenarios, including environmental science, public health, and epidemiological analysis, to build a universal literature-mining platform.
\item Integrate knowledge graphs and interactive visualization to construct evidence map systems for researchers and policy stakeholders.
\end{itemize}

\subsection{Practical Deployment Roadmap}
This work validates the multi-agent literature analysis pipeline on a large-scale real-world bilingual corpus, forming a robust offline prototype for interdisciplinary literature mining. We articulate a clear staged deployment path and address key engineering and sociotechnical challenges toward full real‑world deployment.

First, we will conduct closed pilot testing with domain researchers to optimize system latency, computational cost, and prompt robustness, establishing a human-in-the-loop correction mechanism for stable batch processing. Second, we will release a lightweight public service with accessible analysis APIs and standardized data export functions, supporting open literature analysis for climate and public health communities. Finally, we aim to build a continuously updated platform with automatic literature crawling, incremental knowledge integration, and visualized evidence mapping for long-term academic and policy support.

Our deployment challenges cover both engineering and sociotechnical dimensions. Technically, large-batch processing latency, LLM operational costs, and version management of agent prompts and external knowledge bases require further optimization. From a sociotechnical perspective, automated literature analysis for public health policy demands high result traceability and reliability; inconsistent domain terminology across regional studies requires sustained ontology maintenance. Additionally, human expert supervision remains essential to avoid over-reliance on model-generated results in evidence-based decision making.

Comprehensive offline evaluations in this work provide solid technical foundations for future iterative deployment, and our modular system design enables flexible engineering iteration toward practical, real-world interdisciplinary literature analysis services.

\onecolumn
\newpage
\twocolumn
\section*{Acknowledgments}
This work was supported by the National Natural Science Foundation of China (Grant No. 42175052, U2442206, 42125503) and the China Meteorological Administration Youth Innovation Team (Grant No. CMA2024QN06).

\bibliography{aaai2027}

@article{polak2024extracting,
  title={Extracting accurate materials data from research papers with conversational language models and prompt engineering},
  author={Polak, Maciej P and Morgan, Dane},
  journal={Nature Communications},
  volume={15},
  number={1},
  pages={1569},
  year={2024},
  publisher={Nature Publishing Group UK London}
}

@article{gilligan2023rule,
  title={A rule-free workflow for the automated generation of databases from scientific literature},
  author={Gilligan, Luke PJ and Cobelli, Matteo and Taufour, Valentin and Sanvito, Stefano},
  journal={npj Computational Materials},
  volume={9},
  number={1},
  pages={222},
  year={2023},
  publisher={Nature Publishing Group UK London}
}

@article{charest2026hermes,
  title={HERMES: an open-source mining tool for open-access literature},
  author={Charest, Julien and Priselac, Katarina and Reischer, Georg H and Farnleitner, Andreas H and Mach, Robert L and Mach-Aigner, Astrid R},
  journal={Bioinformatics Advances},
  volume={6},
  number={1},
  pages={vbag058},
  year={2026},
  publisher={Oxford University Press}
}

@article{ke2025large,
  title={Large language models in document intelligence: A comprehensive survey, recent advances, challenges, and future trends},
  author={Ke, Wenjun and Zheng, Yifan and Li, Yining and Xu, Hengyuan and Nie, Dong and Wang, Peng and He, Yao},
  journal={ACM Transactions on Information Systems},
  volume={44},
  number={1},
  pages={1--64},
  year={2025},
  publisher={ACM New York, NY}
}

@article{xu2024large,
  title={Large language models for generative information extraction: A survey},
  author={Xu, Derong and Chen, Wei and Peng, Wenjun and Zhang, Chao and Xu, Tong and Zhao, Xiangyu and Wu, Xian and Zheng, Yefeng and Wang, Yang and Chen, Enhong},
  journal={Frontiers of Computer Science},
  volume={18},
  number={6},
  pages={186357},
  year={2024},
  publisher={Springer}
}

@article{adnan2019analytical,
  title={An analytical study of information extraction from unstructured and multidimensional big data},
  author={Adnan, Kiran and Akbar, Rehan},
  journal={Journal of Big Data},
  volume={6},
  number={1},
  pages={1--38},
  year={2019},
  publisher={Springer}
}

@article{adnan2019limitations,
  title={Limitations of information extraction methods and techniques for heterogeneous unstructured big data},
  author={Adnan, Kiran and Akbar, Rehan},
  journal={International Journal of Engineering Business Management},
  volume={11},
  pages={1847979019890771},
  year={2019},
  publisher={SAGE Publications Sage UK: London, England}
}

@article{dagdelen2024structured,
  title={Structured information extraction from scientific text with large language models},
  author={Dagdelen, John and Dunn, Alexander and Lee, Sanghoon and Walker, Nicholas and Rosen, Andrew S and Ceder, Gerbrand and Persson, Kristin A and Jain, Anubhav},
  journal={Nature communications},
  volume={15},
  number={1},
  pages={1418},
  year={2024},
  publisher={Nature Publishing Group UK London}
}

@inproceedings{sun2023text,
  title={Text classification via large language models},
  author={Sun, Xiaofei and Li, Xiaoya and Li, Jiwei and Wu, Fei and Guo, Shangwei and Zhang, Tianwei and Wang, Guoyin},
  booktitle={Findings of the Association for Computational Linguistics: EMNLP 2023},
  pages={8990--9005},
  year={2023}
}

@inproceedings{zhang2024automated,
  title={Automated mining of structured knowledge from text in the era of large language models},
  author={Zhang, Yunyi and Zhong, Ming and Ouyang, Siru and Jiao, Yizhu and Zhou, Sizhe and Ding, Linyi and Han, Jiawei},
  booktitle={Proceedings of the 30th ACM SIGKDD conference on knowledge discovery and data mining},
  pages={6644--6654},
  year={2024}
}

@inproceedings{yang2023study,
  title={A study on semantic understanding of large language models from the perspective of ambiguity resolution},
  author={Yang, Shuguang and Chen, Feipeng and Yang, Yiming and Zhu, Zude},
  booktitle={Proceedings of the 2023 International Joint Conference on Robotics and Artificial Intelligence},
  pages={165--170},
  year={2023}
}

@article{xu2025toward,
  title={Toward large reasoning models: A survey of reinforced reasoning with large language models},
  author={Xu, Fengli and Hao, Qianyue and Shao, Chenyang and Zong, Zefang and Li, Yu and Wang, Jingwei and Zhang, Yunke and Wang, Jingyi and Lan, Xiaochong and Gong, Jiahui and others},
  journal={Patterns},
  volume={6},
  number={10},
  year={2025},
  publisher={Elsevier}
}

@inproceedings{zeng2024evaluating,
  title={Evaluating large language models at evaluating instruction following},
  author={Zeng, Zhiyuan and Yu, Jiatong and Gao, Tianyu and Meng, Yu and Goyal, Tanya and Chen, Danqi},
  booktitle={International Conference on Learning Representations},
  volume={2024},
  pages={40193--40219},
  year={2024}
}

@article{guo2024large,
  title={Large language model based multi-agents: A survey of progress and challenges},
  author={Guo, Taicheng and Chen, Xiuying and Wang, Yaqi and Chang, Ruidi and Pei, Shichao and Chawla, Nitesh V and Wiest, Olaf and Zhang, Xiangliang},
  journal={arXiv preprint arXiv:2402.01680},
  year={2024}
}

@article{zhang2024comparative,
  title={A comparative study on large language models' accuracy in cross-lingual professional terminology processing: An evaluation across multiple domains},
  author={Zhang, Hanlu and Liu, Wenyan and others},
  journal={Journal of Advanced Computing Systems},
  volume={4},
  number={10},
  pages={55--68},
  year={2024}
}

@inproceedings{kleidermacher2026science,
  title={Science across languages: Assessing LLM multilingual translation of scientific papers},
  author={Kleidermacher, Hannah Calzi and Zou, James},
  booktitle={Findings of the Association for Computational Linguistics: EACL 2026},
  pages={3932--3947},
  year={2026}
}

@article{niu2025flow,
  title={Flow: Modularized agentic workflow automation},
  author={Niu, Boye and Song, Yiliao and Lian, Kai and Shen, Yifan and Yao, Yu and Zhang, Kun and Liu, Tongliang},
  journal={arXiv preprint arXiv:2501.07834},
  year={2025}
}

@inproceedings{qian2025scaling,
  title={Scaling large language model-based multi-agent collaboration},
  author={Qian, Chen and Xie, Zihao and Wang, Yifei and Liu, Wei and Zhu, Kunlun and Xia, Hanchen and Dang, Yufan and Du, Zhuoyun and Chen, Weize and Yang, Cheng and others},
  booktitle={International Conference on Learning Representations},
  volume={2025},
  pages={41488--41505},
  year={2025}
}

@article{kumar2026paper,
  title={Paper Circle: An Open-source Multi-agent Research Discovery and Analysis Framework},
  author={Kumar, Komal and Chadha, Aman and Khan, Salman and Khan, Fahad Shahbaz and Cholakkal, Hisham},
  journal={arXiv preprint arXiv:2604.06170},
  year={2026}
}

@article{xu2025manalyzer,
  title={Manalyzer: End-to-end Automated Meta-analysis with Multi-agent System},
  author={Xu, Wanghan and Zhang, Wenlong and Ling, Fenghua and Fei, Ben and Hu, Yusong and Ma, Runmin and Zhang, Bo and Ren, Fangxuan and Lin, Jintai and Ouyang, Wanli and others},
  journal={arXiv preprint arXiv:2505.20310},
  year={2025}
}

@article{zhang2026researchpilot,
  title={ResearchPilot: A Local-First Multi-Agent System for Literature Synthesis and Related Work Drafting},
  author={Zhang, Peng},
  journal={arXiv preprint arXiv:2603.14629},
  year={2026}
}

@inproceedings{yang2026inex,
  title={InEx: Hallucination Mitigation via Introspection and Cross-Modal Multi-Agent Collaboration},
  author={Yang, Zhongyu and Yuan, Yingfang and Jiang, Xuanming and An, Baoyi and Pang, Wei},
  booktitle={Proceedings of the AAAI Conference on Artificial Intelligence},
  volume={40},
  number={35},
  pages={29829--29837},
  year={2026}
}

@article{zhang2025diting,
  title={DITING: A Multi-Agent Evaluation Framework for Benchmarking Web Novel Translation},
  author={Zhang, Enze and Wang, Jiaying and Xiao, Mengxi and Liu, Jifei and Kuang, Ziyan and Dong, Rui and Dong, Eric and Ananiadou, Sophia and Peng, Min and Xie, Qianqian},
  journal={arXiv preprint arXiv:2510.09116},
  year={2025}
}

@inproceedings{go2026lira,
  title={LiRA: A multi-agent framework for reliable and readable literature review generation},
  author={Go, Gregory Hok Tjoan and Ly, Khang and S{\o}gaard, Anders and Tabatabaei, Seyed Amin and de Rijke, Maarten and Chen, Xinyi},
  booktitle={Proceedings of the AAAI Conference on Artificial Intelligence},
  volume={40},
  number={47},
  pages={40456--40464},
  year={2026}
}

@article{romanello20252025,
  title={The 2025 report of the Lancet Countdown on health and climate change: climate change action offers a lifeline},
  author={Romanello, Marina and Walawender, Maria and Hsu, Shih-Che and Moskeland, Annalyse and Palmeiro-Silva, Yasna and Scamman, Daniel and Smallcombe, James W and Abdullah, Sabah and Ades, Melanie and Al-Maruf, Abdullah and others},
  journal={The Lancet},
  volume={406},
  number={10521},
  pages={2804--2857},
  year={2025},
  publisher={Elsevier}
}

@article{eyzaguirre2025effects,
  title={The effects of climate change on health: a systematic review from a one health perspective},
  author={Eyzaguirre, Indira A Luza and Sousa, Esley Lima de and Martins, Yago de Jesus and Fernandes, Marcus EB and Oliveira-Filho, Aldemir B},
  journal={Climate},
  volume={13},
  number={10},
  pages={204},
  year={2025},
  publisher={MDPI}
}

@article{berrang2021systematic,
  title={Systematic mapping of global research on climate and health: a machine learning review},
  author={Berrang-Ford, Lea and Sietsma, Anne J and Callaghan, Max and Minx, Jan C and Scheelbeek, Pauline FD and Haddaway, Neal R and Haines, Andy and Dangour, Alan D},
  journal={The Lancet Planetary Health},
  volume={5},
  number={8},
  pages={e514--e525},
  year={2021},
  publisher={Elsevier}
}

@article{lee2026explainable,
  title={An Explainable Approach to Document-level Translation Evaluation with Topic Modeling},
  author={Lee, Hyeokmin and Kim, Youngkyu and Yoo, Byounghyun},
  journal={arXiv preprint arXiv:2604.20334},
  year={2026}
}

@article{berrang2021mapping,
  title={Mapping global research on climate and health using machine learning (a systematic evidence map)},
  author={Berrang-Ford, Lea and Sietsma, Anne J and Callaghan, Max and Minx, Ja C and Scheelbeek, Pauline and Haddaway, Neal R and Haines, Andy and Belesova, Kristine and Dangour, Alan D},
  journal={Wellcome open research},
  volume={6},
  pages={7},
  year={2021}
}

@article{zander2023topic,
  title={Topic modelling the mobility response to heat and drought},
  author={Zander, Kerstin K and Baggen, Hunter S and Garnett, Stephen T},
  journal={Climatic Change},
  volume={176},
  number={4},
  pages={42},
  year={2023},
  publisher={Springer}
}

@article{nasari2026healthy,
  title={A healthy cities agenda for extreme heat adaptation in urban settings},
  author={Nasari, Alaha and Marzouk, Sammer},
  journal={Bulletin of the World Health Organization},
  volume={104},
  number={3},
  pages={163},
  year={2026}
}

@misc{openalex,
  title        = {OpenAlex: A fully-open index of scholarly works, authors, venues, institutions, and concepts},
  author       = {Heather Piwowar and Jason Priem and Richard Orr},
  year         = {2022},
  howpublished = {\url{https://openalex.org}},
  note         = {Accessed: 2026-05-22},
  doi          = {10.48550/arXiv.2205.01833}
}

@misc{cnki,
  title        = {China National Knowledge Infrastructure (CNKI)},
  author       = {Tsinghua University},
  year         = {1999},
  howpublished = {\url{https://www.cnki.net}},
  note         = {Accessed: 2026-05-22},
  language     = {Chinese}
}

@misc{baiduscholar,
  title        = {Baidu Scholar},
  author       = {Baidu Inc.},
  year         = {2014},
  howpublished = {\url{https://xueshu.baidu.com}},
  note         = {Accessed: 2026-05-22},
  language     = {Chinese}
}

@misc{wmo_metelib,
  title        = {WMO Meteorological Element Library},
  author       = {World Meteorological Organization (WMO)},
  year         = {2025},
  howpublished = {\url{https://codes.wmo.int/}},
  note         = {Accessed: 2026-05-22}
}

@misc{cn_placelib,
  title        = {Chinese City Place Name Library},
  author       = {Ministry of Civil Affairs of China},
  year         = {2020},
  howpublished = {\url{https://www.mca.gov.cn}},
  note         = {Accessed: 2026-05-22},
  language     = {Chinese}
}

@misc{who_icd,
  title        = {International Classification of Diseases (ICD-11)},
  author       = {World Health Organization (WHO)},
  year         = {2022},
  howpublished = {\url{https://icd.who.int/en/}},
  note         = {Accessed: 2026-05-22}
}

@article{cai20252025,
  title={The 2025 China report of the Lancet Countdown on health and climate change: empowering cities for synergistic action},
  author={Cai, Wenjia and Zhang, Chi and Zhang, Shihui and Bai, Yuqi and Chen, Bin and Chen, Jifei and Cheng, Liangliang and Fan, Weicheng and Feng, Lianfang and Guan, Dabo and others},
  journal={The Lancet Public Health},
  volume={10},
  number={12},
  pages={e1066--e1085},
  year={2025},
  publisher={Elsevier}
}

@inproceedings{devlin2019bert,
  title={Bert: Pre-training of deep bidirectional transformers for language understanding},
  author={Devlin, Jacob and Chang, Ming-Wei and Lee, Kenton and Toutanova, Kristina},
  booktitle={Proceedings of the 2019 conference of the North American chapter of the association for computational linguistics: human language technologies, volume 1 (long and short papers)},
  pages={4171--4186},
  year={2019}
}

@article{wu2023autogen,
  title={Autogen: Enabling next-gen llm applications via multi-agent conversation},
  author={Wu, Qingyun and Bansal, Gagan and Zhang, Jieyu and Wu, Yiran and Li, Beibin and Zhu, Erkang and Jiang, Li and Zhang, Xiaoyun and Zhang, Shaokun and Liu, Jiale and others},
  journal={arXiv preprint arXiv:2308.08155},
  year={2023}
}


\newpage

\onecolumn
\section{Appendix}

\renewcommand{\thefigure}{S\arabic{figure}} 
\renewcommand{\thetable}{S\arabic{table}} 
\setcounter{figure}{0} 
\setcounter{table}{0}  
\subsection{Table}

\begin{table*}[h]
\centering
\caption{Ablation on hallucination‑control components and central coordination module. Mean $\pm$ std over three runs.}
\label{tab:ablation_hallucination}
\begin{tabular}{lccc}
\toprule
Variant & Precision & Recall & F1‑score \\
\midrule
Ours (full system) & 0.93$\pm$0.01 & 0.92$\pm$0.01 & 0.92$\pm$0.01 \\
w/o central coordination module & 0.91$\pm$0.02 & 0.89$\pm$0.02 & 0.90$\pm$0.02 \\
w/o cross‑agent verification & 0.89$\pm$0.02 & 0.88$\pm$0.01 & 0.88$\pm$0.02 \\
w/o confidence‑score filtering & 0.87$\pm$0.01 & 0.90$\pm$0.02 & 0.88$\pm$0.01 \\
w/o external knowledge‑base filtering & 0.88$\pm$0.02 & 0.87$\pm$0.02 & 0.87$\pm$0.02 \\
\bottomrule
\end{tabular}
\end{table*}


\subsection{Experiment Details}
\subsubsection{Evaluation Metrics}
The framework performance is evaluated from four dimensions: information extraction accuracy, system operation efficiency, multilingual processing performance, and domain application value to ensure comprehensive and reliable results.

\subsubsection{Information Extraction Accuracy Metrics}
Taking manual annotation results of two climate-health domain experts as the gold standard, precision (P), recall (R), and F1-score (F1) of the framework in core information extraction tasks are calculated, with F1-score as the core evaluation metric balancing precision and recall. The corresponding formulas are:

\begin{equation}
P = \frac{TP}{TP + FP} ,
\end{equation}

\begin{equation}
R = \frac{TP}{TP + FN} ,
\end{equation}

\begin{equation}
F1 = 2 \times \frac{P \times R}{P + R} ,
\end{equation}

\noindent where TP is the number of correctly-extracted information entities, FP is the number of hallucinatory information incorrectly extracted, and FN is the number of valid information not extracted. Core extraction tasks include geographic information (city), climate exposure factor, and health outcome, with metrics calculated for each task and the overall average taken.

\subsubsection{Experimental Data and Corpus Construction}
To comprehensively validate the effectiveness of the framework, we constructed a Chinese-English bilingual climate-health literature corpus covering mainland China from 1993 to 2023, sourced from authoritative academic databases to ensure the representativeness and academic quality of the corpus.

\subsubsection{Data Sources and Retrieval Strategy}
The corpus is mainly derived from three databases:

\begin{itemize}
\item {\texttt{OpenAlex}}: Used to obtain English international journal literature, with a retrieval time range of 1993–2023, and 58,148 initial retrieved papers covering mainstream international research in the climate-health field.

\item {\texttt{China National Knowledge Infrastructure (CNKI)}}: Used to obtain Chinese core journal literature, with a retrieval time range of 2009–2023, and 2,561 initial retrieved papers collecting high-quality domestic climate-health research.

\item {\texttt{Baidu Scholar}}: As a supplementary source for Chinese literature, with a retrieval time range of 2009–2023, and 2,731 initial retrieved papers further improving domestic research coverage.
\end{itemize}

Retrieval keywords adopt the combination mode of "climate/\-meteorological factor + health outcome" in both Chinese and English versions, with retrieval scope limited to title, abstract, and keyword fields to ensure maximum recall of relevant literature. To ensure corpus quality, only literature published in high-quality journals such as SCI/EI, Peking University Core, CSCD, and CSSCI are included, filtering out low-quality outputs other than conference abstracts and reviews.
  
\subsubsection{Data Preprocessing and Corpus Scale}
The raw corpus undergoes a standardized preprocessing process to remove noise and duplicate data, forming the final experimental corpus:

\begin{itemize}
\item {\texttt{Deduplication}}: Duplicate published literature is eliminated based on fuzzy matching of title, author, and publication year (similarity threshold set to 0.95) to avoid repeated analysis.
\item {\texttt{Metadata Standardization}}: Metadata formats of different databases are unified, journal name, publication year, author information and other fields are standardized, and missing core metadata are supplemented.
\item {\texttt{Text Structuring}}: Title, abstract, and keywords of each paper are extracted, non-text information such as formulas, charts, and tables is removed, converted to plain text format, and literature language type (Chinese/English) is labeled.
\end{itemize}

After preprocessing, 32,642 valid papers are finally obtained, including 28,150 English papers and 4,492 Chinese papers, forming the basic corpus for experiments.

\subsection{Experimental Setup and Implementation Details}

\subsubsection{Multi-Agent Framework Configuration}
The framework is developed based on Python 3.9 and adopts a centralized scheduling architecture. The central coordination module is implemented based on the Prefect workflow engine, responsible for task distribution and data flow. The three major agents (document appraiser, information extractor, senior analyst) are connected to mainstream LLM foundation models (Deepseek, ChatGPT, Ernie Bot), with temperature parameter set to 0.1 to reduce random output and ensure result consistency and reproducibility.

\subsubsection{External Support Resource Configuration}
To improve domain adaptability and accuracy, the system integrates three types of external databases:

\begin{itemize}
\item {\texttt{WMO Meteorological and Climate Element Library}}: Provides standardized climate variable terminology and classification system to assist agents in identifying and verifying climate exposure factors.
\item {\texttt{Chinese City Place Name Database}}: Contains standard names and administrative division information of all cities in China for geographic information verification and standardization.
\item {\texttt{WHO-ICD Disease and Health Factor Library}}: Provides standardized health outcome terminology and classification to ensure the standardization of health outcome extraction.
\end{itemize}

\subsubsection{Gold‑standard Annotation Setup}
We randomly sampled $N_{sample}=800$ papers (mixed Chinese and English) as the held‑out annotated gold corpus. Two climate‑health domain experts independently annotated core extraction fields (geographic location, climate exposure factor, health outcome). Inter‑Annotator Agreement (IAA) is computed using macro‑F1 over structured extraction outputs, reaching $0.87$. Disagreements are resolved via joint discussion. Experts do not annotate the full 32,642 corpus; all automatic evaluation metrics are computed on this 800‑paper annotated subset.

\subsubsection{System Efficiency Metrics}
Processing time and throughput of the framework and manual analysis are compared to evaluate efficiency improvement, with specific metrics including:

\begin{itemize}
\item {\texttt{Processing time}}: Time required to process the same number of literature (hours).
\item {\texttt{Throughput}}: Number of literature processed per unit time (papers/hour).
\item {\texttt{Efficiency improvement ratio}}: Ratio of manual analysis throughput to framework throughput.
\end{itemize}

\subsubsection{Multilingual Processing Performance Metrics}
Extraction accuracy (F1-score) of the framework on Chinese and English literature is calculated separately to evaluate cross-language processing capability.

\subsubsection{Domain Application Value Metrics}
From a practical application perspective, the usability of the framework output dataset is evaluated, with metrics including:

\begin{itemize}
\item {\texttt{Effective extraction rate}}: Proportion of literature with successfully extracted core information in relevant literature.
\item {\texttt{Data standardization rate}}: Proportion of extraction results matching the domain classification framework.
\item {\texttt{Research gap identification ability}}: Ability to identify spatial and thematic gaps in domain research based on the output dataset.
\end{itemize}

\subsection{Agent System Prompts}
Representative system prompts for three agents are provided below. The full prompt set and configuration files are released in our open repository.

\subsubsection{Document Appraiser Agent Prompt}
\begin{quote}
You are a document relevance screener for climate-health literature.
Given title, abstract, keywords, judge whether this paper studies climate-related exposure and health outcomes in Chinese mainland cities.
Only output binary result: relevant / irrelevant.
Do not make extra inference; judge only based on explicit text content.
\end{quote}

\subsubsection{Information Extractor Agent Prompt}
\begin{quote}
Extract structured information from given paper text.
Allowed fields: research city, climate exposure factor, health outcome, vulnerable population, socioeconomic factors.
Only extract entities explicitly mentioned in original text. Do NOT infer or fabricate facts.
For each extracted entity, output a confidence score from 1 to 5.
For multiple entities, split into atomic independent records.
Do not arbitrarily combine entities from different text fragments.
Output structured JSON format.
\end{quote}

\subsubsection{Senior Analyst Agent Prompt}
\begin{quote}
Verify and normalize extracted records.
Invoke external knowledge lookup tools to validate city names, climate terms and health outcomes.
Filter records with confidence score lower than 3.
Check logical consistency between climate exposure and health outcome.
Generate analysis statistics based on validated records.
\end{quote}

\end{document}